\documentclass[letterpaper, 10 pt, article]{ieeeconf}  

\IEEEoverridecommandlockouts                              

\usepackage[
    style=ieee,
    doi=false,
    isbn=false,
    url=false,
    eprint=false,
    backend=bibtex,
    natbib=true
    ]{biblatex}
\usepackage{csquotes}
\usepackage{graphicx}
\usepackage{subcaption}
\usepackage{amsmath}
\usepackage{blindtext}

\renewcommand{\cite}{\parencite}

\usepackage{amsmath}
\usepackage{amsfonts}

\usepackage{perl_acronyms}
\usepackage{perl_misc}
\newcommand{\subparagraph}{}

\usepackage{filecontents}
\usepackage{silence}
\usepackage{graphics} 
\usepackage{tikz}
\usetikzlibrary{shapes.geometric, arrows, calc, positioning, fit}
\tikzstyle{startstop} = [ellipse, minimum width=1cm, minimum height=1cm, text centered, draw=black, fill=white]

\tikzstyle{sum} = [circle, minimum width=0.15cm, minimum height=0.15cm, text centered, draw=black, fill=white]

\tikzstyle{ctrl} = [rectangle, minimum width=2.5cm, minimum height=1.25cm,text centered, draw=black, fill=white]

\tikzstyle{arrow} = [thick,->,>=stealth]
\tikzset{line/.style={draw, thick, -latex'}}

\title{\LARGE \bf
A constrained control-planning strategy for redundant manipulators 
}

\author{Corina Barbalata$^{1}$, Ram Vasudevan$^{2}$ and Matthew Johnson-Roberson$^{1}$  \vspace{-2ex}
\thanks{$^{1}$Corina Barbalata and Matthew Johnson-Roberson are with the Department of Naval Architecture and Marine Engineering, University of Michigan, United States; $^{2}$Ram Vasudevan is with the Department of Mechanical Engineering, University of Michigan, United States
{\tt\small \{corinaba,ramv,mattjr\}@umich.edu}}
}
\begin{document}

\maketitle

\textbf{\textit{Abstract} -
This paper presents an interconnected control-planning strategy for redundant manipulators, subject to system and environmental constraints. 
The method incorporates low-level control characteristics and high-level planning components into a robust strategy for manipulators acting in complex environments, subject to joint limits.
This strategy is formulated using an adaptive control rule, the estimated dynamic model of the robotic system and the nullspace of the linearized constraints. A path is generated that takes into account the capabilities of the platform. 
%
The proposed method is computationally efficient, enabling its implementation on a real multi-body robotic system. 
Through experimental results with a $7$ \ac{DOF} manipulator, we demonstrate the performance of the method in real-world scenarios.  
}

\section{Introduction}
\label{sec:introduction}
\begin{figure}[!htb]
  \centering
    \includegraphics[width=0.85\columnwidth]{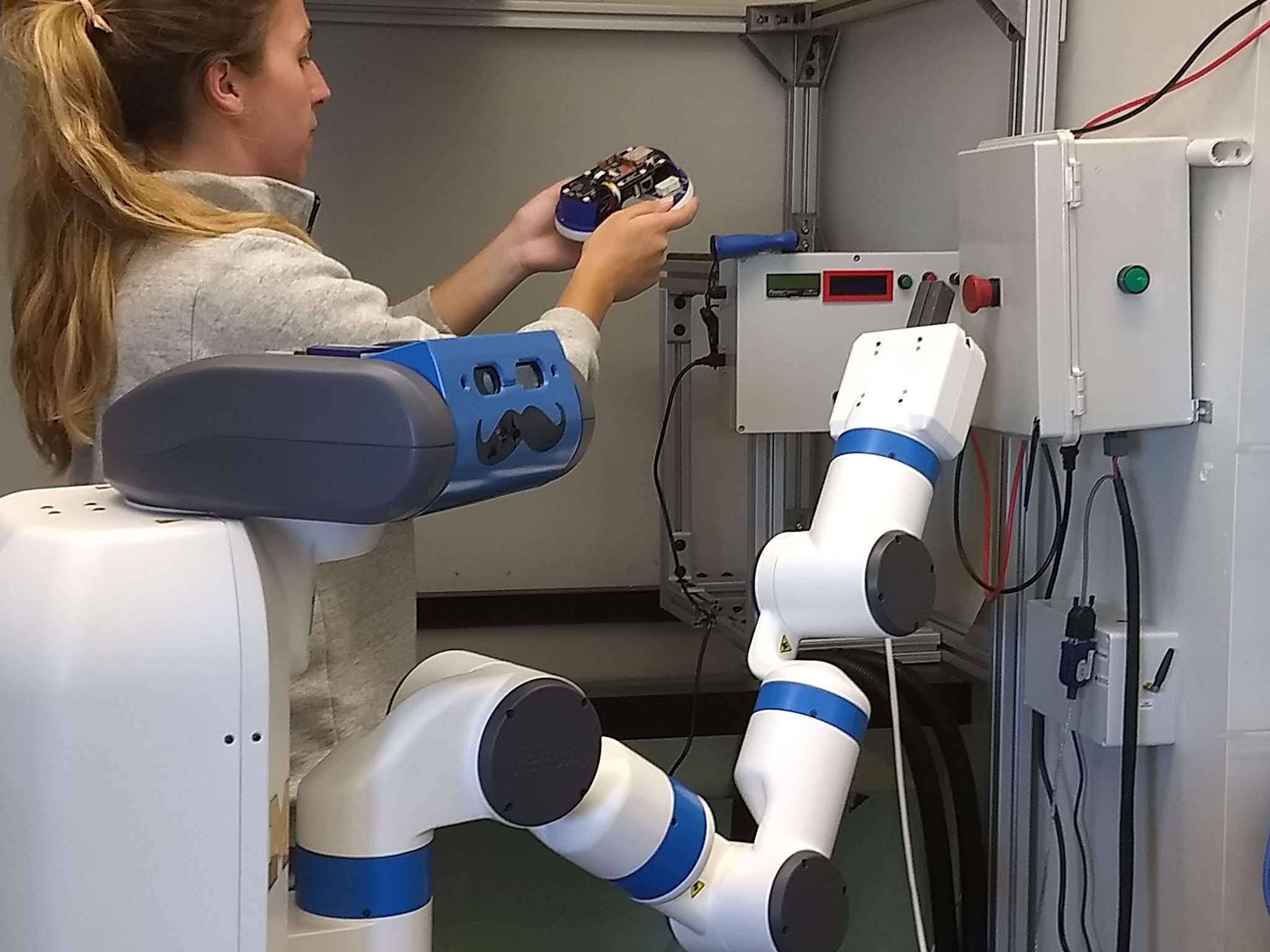}
     \caption{\small{Robot working in a cluttered environment in close proximity to people: the Fetch robot \cite{wise2016fetch} reaching for the emergency button, ensuring minimum joint movement to not harm humans in a collaborative environment.}}
     \label{fig:fetch}
\end{figure}

Recent advances in \ac{AI} and improvements in hardware have led to the development of sophisticated robotic systems. 
Industrial robots  performing tasks on factory production lines since the $1970$s have evolved to be extremely dexterous, have increased learning capabilities and precise movements  \cite{zhang2017development}. 
The newer types of robots with humanoid shapes or more practical configurations are  mobile and designed to perform tasks in cluttered and dynamic environments, which demand robust and safe behavior. 
Unfortunately due to their complex mechnanical design, nonlinear characteristics and limited actuation capabilities, the mobile robots have difficulty performing complex tasks in restrictive environments.
Some of these limitations can lead to issues in the motion planning component of the robotic system, in some cases preventing the planner to avoid collisions or to produce efficient plans to enable real-time performance. 
%
In addition to high-level motion planning, limitations of low-level controllers hinder robot performance, either producing unsafe motions through their attempt to achieve the plans handed to them or failing to achieve their goals while remaining safe. 
Thus, robust low-level strategies together with a detailed knowledge of the system can be used as an initial estimation for obtaining feasible plans for the desired task. 
Mobile robotic systems that act in cluttered spaces have to not only fulfill a single task, but they also have to compensate for the restrictions imposed by the environment.
As presented in \cite{nakamura1987task} these constraints can be formulated as a series of sub-tasks that the robot must  fulfill to ensure a robust, safe, and accurate action plan. 
For instance, in a scenario where robots have to work in close proximity to people in narrow environments, as depicted in Figure~\ref{fig:fetch}, the sub-tasks that would result in a successful execution are: \textit{(i)} reaching the goal (emergency button in this case);  \textit{(ii)} guaranteeing that the arm does not hurt the people working in its surroundings or collide with any object; \textit{(iii)} ensure that the physical limitations (joint limits) 
of the robot are not broken. 
A successful outcome of this scenario is when all these sub-tasks are completed without conflicting with one another. It is essential to highlight that although reaching the goal is important, this goal has lower priority than safe operation. This results in creating solutions based on a priority policy \cite{nakamura1987task}. 
Each task has a strict priority and it is only achieved when its execution does not violate a higher priority task.  
\vspace{-2pt}

\vspace{-10pt}
In the context of a complex robotic system acting in dynamic and cluttered environments, planning strategies that ensure safe manipulation are essential.
%
With this paper we address the problem of manipulation in cluttered environments by proposing the following contributions: $(1)$ A strategy for producing a robust initial estimate for trajectory generation subject to constraints; $(2)$ The addition of inequality constraints in a prioritized-optimal control formulation; and $(3)$ Experimental evaluation of the proposed strategy with the $7$ \ac{DOF} manipulator of a Fetch robot, acting in a restrictive environment, such as the one seen in Figure~\ref{fig:fetch}.

This paper is structured as follows: In Section \ref{sec:background}, we provide background on state-of-the-art control and planning methods for complex systems. 
Section \ref{sec:prob_statemeny} presents the problem using an optimal control formulation. Section \ref{sec:proposed_method} presents the proposed algorithm. 
Section \ref{sec:results} discusses the experimental results, and Section \ref{sec:conclusions} summarizes the paper.

\vspace{-10pt}
\section{Background}
\label{sec:background}

%
Optimal control strategies rely on an initial low-level control structure to obtain feasible high-level solutions  \cite{kirk2012optimal}. 
These methods are advantageous because they take into account future behavior of the robotic system, but have limitations in handling multiple tasks that are common for manipulation in cluttered environments. 
%
An alternative approach in handling multiple tasks for robotic manipulation is prioritized control \cite{nakamura1987task}. 
This approach offers a way to compute the control law based on the importance (i.e., priority) of each task that the robot has to fulfill. 
%
%
Used for velocity control for a $7$ \ac{DOF} manipulator in  \cite{slotine1991general} the prioritized control method was extended for  dynamic control in \cite{sentis2007synthesis} and used in a humanoid simulation environment. 
Additionally, a computational fast inverse dynamics priority control technique was proposed in \cite{righetti2011inverse} and its validity was demonstrated by numerical evaluation on a floating base robotic system. 
The issue with the classical priority control formulation for robotic manipulators is their inability to handle inequality constraints, such as joint limit constraints. Furthermore, obstacle avoidance problems can be represented in configuration space as inequality constraints, priority control methods failing to handle these cases. 
%
A well known approach for handling these inequality constraints is based on an artificial potential function \cite{khatib1986potential}, defined as a virtual force driving the manipulator away from the obstacle or the joint limit \cite{marchand1998dynamic}. 
Another efficient priority control method that includes manipulator joint limits and motor torque inequality constraints was proposed in \cite{saab2011generation}. 
In this case, the authors used slack variables to make the transformation from inequality to equality constraints. 
Handling inequality constraints was also proposed in \cite{antonelli2009prioritized} for a $6$ \ac{DOF} planner manipulator, where the solution to the unbounded problem was computed and then iteratively restricted within the boundaries. 
\ac{QP} approaches for kinematic priority control have been proposed in \cite{kanoun2009prioritizing}, \cite{kanoun2012real} for planning local motions for a humanoid robot. 
For each task, they create a quadratic program with a weight that defines its priority. The constraints of the problem were handled by the numerical solver, although for multiple tasks the weight adjustment can be challenging. 
In \cite{betts1993path} \ac{SQP} was defined, which proposed through system discretization that the optimal control problem could be formulated into a nonlinear problem having a direct solution. 
This method is considered one of the most attractive strategies to obtain numerical solutions in the case of nonlinear optimal problems~\cite{fiacco1990nonlinear}, and this formed the foundation of priority optimal control for mobile manipulation.

A strategy for control and planning problems that incorporates the benefits of both optimal and prioritized control is the prioritized-optimal architecture \cite{del2014prioritized} where a sequence of objectives are optimized based on the importance of the task.
The task prioritization for optimal control was proposed using an initial control strategy and a cascade of \ac{QP} problems. 
The strategy was applied on a simulated humanoid robot and aims to solve the problem of tuning the task weights. 
In an effort to reduce the cubic complexity of the problem, \cite{romano2015prioritized} introduced a hierarchical model into the dynamic programming, which is capable of handling task priorities. 
A prioritized-optimal control with linear time-complexity in the number of time steps was presented in \cite{giftthaler2017projection}.
The method is based on an iterative \ac{LQR} algorithm and nullspace projection method to handle equality constraints. 
However, the method overlooks the initial control strategy, a \ac{LQR} feedback controller, and its importance in obtaining accurate solutions, it does not handle inequality constraints and has high computation times, preventing the current prioritized optimal control methods to be applied on high-dimensional mobile robots in real time.
With this work, we aim to address these limitations and ensure that the generated path is always feasible from a low-level control perspective and satisfies all dynamic characteristics of the robot.


\vspace{-10pt}
\section{Problem formulation}
\label{sec:prob_statemeny}
This paper studies the problem of a multi-dimensional robotic manipulator performing tasks in restrictive environments from a control and planning perspective.  Let us define a nonlinear optimal control problem subject to inequality constraints as:
\begin{equation}
     \min_{u_i} \left \{ \phi(x_N) + \sum_{i=0}^{N-1} f_i(x_i,u_i,i)     \right\}
      \label{eq:problem_statement}
  \end{equation}
  \vspace{-2pt}
subject to:
\vspace{-2pt}
\begin{equation}
    x_{i+1} = g(x_i,u_i,i), \quad x(0) =x_0
    \label{eq:dynamic_eq}
\end{equation}
\begin{equation}
    h_1(x_i, i) \leq 0
    \label{eq:constraint_1}
\end{equation}
\begin{equation}
    h_2(x_i, i) \leq 0
    \label{eq:constraint_2}
\end{equation}
 where $x_i \in \mathbb{R}^n$ is the state vector consisting of joint positions, $u_i \in \mathbb{R}^m$ is the control vector consisting of joint velocities, $f_i$ is the non-negative cost function, $\phi$ is the terminal cost function, $h_1$ and $h_2$ are the non-linear inequality constraints, and $x(0) = x_0$ represent the initial conditions. For this problem, we propose: \textbf{\textit{(i)}} a robust  estimation method to be used as an initial guess in the optimal control formulation and \textbf{\textit{(ii)}} a real-time feedback solution that incorporates inequality constraints.

\vspace{-5pt}
\section{Proposed method}
\label{sec:proposed_method}
\tikzstyle{block} = [draw, fill=white, rectangle, 
    minimum height=3em, minimum width=6em]
\tikzstyle{sum} = [draw, fill=white, circle, node distance=1cm]
\tikzstyle{input} = [coordinate]
\tikzstyle{output} = [coordinate]
\tikzstyle{pinstyle} = [pin edge={to-,thin,black}]

In this section we present a path generation algorithm for redundant manipulation systems. A new method for generating a robust initial path estimation is described in  Section~\ref{sub:lla_control} and in Section \ref{sub:optimal} we incorporate this initial estimate together with inequality constraints into a prioritized-optimal control strategy.
\vspace{-10pt}
\subsection{Adaptive path estimation}
\label{sub:lla_control}

We propose a strategy for defining a robust initial guess for optimal control strategies, that can be divided into two stages: $(a)$ design of a low-level control law and $(b)$ estimation of the initial path. 
\paragraph{Low-level control law}
In the first step the low-level control forces are computed using a strategy that requires one single parameter to be tuned and ensures robust behavior in the presence of system uncertainties. 
The controller has a feedback structure and has the capability to provide a reliable transient response. 
It is characterized by a single control loop for position control, Equation~\eqref{eq:feed_forward_adaptive_pid}, defined by the joint error, ${e_{x_i} = x_{goal}  - x_i}$.
The position goal, $x_{goal} \in \mathbb{R}^n$, represents the final joints desired configuration.
Using an iterative formulation the feedback control law, with gravity compensation, is defined as:
\begin{equation}
    \tau_i =  \gamma \Lambda(e_{{x}_i}) + \eta
    \label{eq:feed_forward_adaptive_pid}
\end{equation}
where $\gamma \in \mathbb{R}^n$ is the adaptive gain parameter that describes the rate of change of the controller, $\tau_i \in \mathbb{R}^n$ is the control input for all the joints of the manipulator, at time-step $i$ and $\Lambda$ is a nonlinear function dependent on the joint errors defined:  
\vspace{-10pt}
\begin{equation}
\begin{aligned}
\Lambda(e_{{x}_i}) =  e_{{x}_i}^3 + e_{{x}_i} \left( \frac{e_{{x}_i} - e_{{x}_{i-1}}}{\Delta t}\right) ^2  +  e_{{x}_i}  \left( \sum_{j=0}^i  e_{{x}_j} \Delta t \right)^2 
\end{aligned}
    \label{eq:control_function}
\end{equation}
Similar to a \ac{PID} controller Equation~(\ref{eq:control_function}) aims to eliminate residual error and estimate the future behaviour of the system. 
By using a formulation that is proportional, and of equal sign to the change in the control error, the proposed  method is able to drive the system to steady-state. 
Furthermore, it is only dependent on a single parameter that characterizes the rate of change in the controller. 

\paragraph{Path estimation}
In this stage we use the low-level control law defined in the previous step with an estimated dynamic model to generate the initial path estimation.
\begin{equation}
\begin{aligned}
\hat{x}_{(i+1), {des}} = & Ax_i +B\tau_i \\
\hat{u}_{i, {des}} = & Gx_i
\end{aligned}
\label{eq:estimated_path}
\end{equation}
where $x_{(i+1), {des}} \in \mathbb{R}^n$ represents the initial estimate path, $A \in \mathbb{R}^{n \times n}$, $B \in \mathbb{R}^{n \times m}$, $G \in \mathbb{R}^{m \times n}$ represent the estimated system, input and output matrices obtained from the dynamic model of the platform. Due to uncertainties in the mathematical model of the robot these values are approximated.
\vspace{-10pt}
\subsection{Optimal-Adaptive path generation}
\label{sub:optimal}
The estimated path defined in Equation~(\ref{eq:estimated_path}) takes into account the torque limitations but does not incorporate environmental constraints or the physical limitations of the system. 
To address these we use a null-space constrained optimal formulation, Equation~(\ref{eq:problem_statement}), to obtain solutions for a feasible path. 
This approach has been used for prioritized-optimal control previously in \cite{de2009prioritized} and \cite{romano2015prioritized} for low-level control structures. 
We extend the work presented in \cite{giftthaler2017projection} by proposing a solution that handles inequality constraints. 
 \subsubsection{Inequality constraints}

In this section we introduce the inequality constraints to the control problem.
We use the artificial potential field approach as an effective real-time method for incorporating constraints. 
This approach has been demonstrated for state and state-input constraints such as joint limits and obstacle avoidance \cite{khatib1986real}.
 
State constraints can be expressed as Equation (\ref{eq:joint_limits_min}), if a \textit{minimum} problem is described, or as Equation (\ref{eq:joint_limits_max}) for a \textit{maximum} problem:
\begin{equation}
    h_2(x_i, i) = \begin{cases}
     \eta \left( \frac{1}{\underline{\rho}_i} - \frac{1}{\underline{\rho}_0} \right) \frac{1}{\underline{\rho}_i^2}, & \quad \text{if $\underline{\rho}_i \leq \underline{\rho}_0$}    \\ \!
      0, & \quad \text{if $\underline{\rho}_i > \underline{\rho}_0$}
    \end{cases}
    \label{eq:joint_limits_min}
\end{equation}
\vspace{-10pt}
\begin{equation}
    h_3(x_i, i) = \begin{cases}
     - \eta \left( \frac{1}{\bar{\rho}_i} - \frac{1}{\bar{\rho}_0} \right) \frac{1}{\bar{\rho}_i^2}, & \quad \text{if $\bar{\rho}_i \leq \bar{\rho}_0$}    \\ \!
      0, & \quad \text{if $\bar{\rho}_i > \bar{\rho}_0$}
    \end{cases}
    \label{eq:joint_limits_max}
\end{equation}

where $\underline{\rho}_0$ and $\bar{\rho}_0$ represent the distance limit to the potential field, $\underline{\rho}_i = x_i - \underline{x}_i$ is the distance from the minimum limit, and $\bar{\rho}_i = \bar{x}_i - x_i$ is the distance from the maximum limit. As presented by \citet{khatib1986potential} this method of representing system limitations and obstacle avoidance is an attractive approach, as it provides the global information necessary for robot control applications. 
\subsubsection{Optimal framework formulation}
Following work done in \cite{li2004iterative}, \cite{murray1984differential} and \cite{del2014prioritized}, the Newton's method for nonlinear minimization is used to reformulate the optimal formulation presented in Equation (\ref{eq:problem_statement}). 
A new linear-quadratic constrained problem that describes the effects of perturbations on the state variables (joint positions) and input variables (joint velocities) is defined, allowing an iterative formulation of the initial optimization problem. The displacement from the measured state, $\Delta x_i = x_i - \hat{x}_{i,{des}}$ and from the input state $\Delta u_i = u_i - \hat{u}_{i,{des}}$ are expressed based on the estimated desired joint positions and velocities in Equation~(\ref{eq:estimated_path}).

Using these displacements, the inequality constraints presented in Equations~(\ref{eq:joint_limits_min})-(\ref{eq:joint_limits_max}), the optimal control problem expressed by Equations~ (\ref{eq:problem_statement})-(\ref{eq:constraint_2}) can now be reformulated dependent on the state and input deviations as a linear quadratic control problem:
 \begin{equation}
 \begin{aligned}
    \Delta u^{*}_i =  \min_{\Delta u_i} &\lbrace  o_N + \Delta x_N^T q_N + \frac{1}{2} \Delta x_N^T Q_N + \Delta x_N  + \\ \!
    & + \sum_{i=0}^{N-1} o_i + \Delta x_i q_i + \Delta u_i^T r_i + \frac{1}{2} \Delta x_i^T Q_i \Delta x_i +  \\ \!
    & + \frac{1}{2} \Delta u_i^T R_i \Delta u_i + \Delta u_i^T P_i \Delta x_i  \rbrace
\end{aligned}
      \label{eq:problem_statement_final}
  \end{equation}
  subject to:
 \begin{equation}
    \Delta x_{i+1} = A_i \Delta x_i + B_i \Delta u_i, \quad x(0) =x_0
    \label{eq:dynamic_eq_final}
\end{equation}
\begin{equation}
    D_i \Delta x_i  = e_i 
    \label{eq:constraint_1_final}
\end{equation}
\begin{equation}
    C_i \Delta x_i = d_i
    \label{eq:constraint_2_final}
\end{equation}

where $o, \ q, \ r, \ Q, \ R, \ P $ are coefficients of the Taylor expansion of Equation (\ref{eq:problem_statement}) around the estimated states. 
The linearized dynamics are presented in Equation (\ref{eq:dynamic_eq_final}) and the inequality constraints are re-defined using the potential fields in  Equations (\ref{eq:constraint_1_final}) - (\ref{eq:constraint_2_final}). To handle any possible discontinuities in the formulation of the constraints we used the method presented in \cite{stella2016very}.
The goal is to solve this local-optimal control law generating a planning policy update, $\Delta u_i(x_i)$, which will be used in the global formulation:
\begin{equation}
    u_i(x_i) = u_i(x_i) + \Delta u_i(x_i)
    \label{eq:final_ctrl_law}
\end{equation}
This planning policy update is responsible in refining the initial planning strategy defined by Equation~(\ref{eq:estimated_path}), taking into account restrictions caused by the working environment and generating an optimal trajectory.  

Following the work of \citet{giftthaler2017projection}, we use the nullspace projection theory to present a solution for the optimal formulation in Equation (\ref{eq:problem_statement_final}), by:
\begin{equation}
   \Delta u_i = \Delta u_i^c + \mathcal{N}\left( \begin{bmatrix}
    E_i\\
    M_i
    \end{bmatrix} \right) \Delta u_i^m
    \label{eq:null_space_reform}
\end{equation}
where $\Delta u_i^c$ is the constrained  control component that ensures the constraints are always fulfilled, $E_i = D_{i+1}B_i$ and $M_i = C_{i+1} B_i$ are the propagation of the input displacements to the next step of the constraints and $\mathcal{N}$ is the projection of the nullspace of these linearized constraints. $\Delta u_i^m$ is the control component that will ensure a valid solution to the optimal problem without breaking the constraints.

The constrained control law, $\Delta u_{i}^c$, is defined to satisfy the dynamics of the constraints at the next step in time. To achieve this we propose the use of feedforward-feedback strategy:
\begin{equation}
     \Delta u_{i}^c =  \begin{bmatrix}
    E_i\\
    M_i
    \end{bmatrix}^{\dagger} 
      \begin{bmatrix}
    e_{i+1}\\
    d_{i+1}
    \end{bmatrix} -  \begin{bmatrix}
    E_i\\
    M_i
    \end{bmatrix}^{\dagger}  \begin{bmatrix}
    F_i\\
    N_i
    \end{bmatrix} (\Delta x_{i_{des}} + \alpha)
    \label{eq:feed_forward}
\end{equation}
where $ \left[ E_i^T, \ M_i^T \right]^{\dagger} $ is the unweighted Moore-Penrose pseudoinverse matrix. $F_i = D_{i+1}B_i$ and $N_i = C_{i+1}B_i$ are the propagation of the state displacements to the next step of the constraints. To compensate for the disturbances in the actuators and inaccuracies in the robot unmodeled dynamics (damping and friction), we propose to introduce a feedback law, $\alpha \in \mathbb{R}^6$,  to make the control scheme robust. The proposed feedback law is of the same type as the one presented in Section \ref{sub:lla_control}, in this case being directly dependent of the difference between the estimated state, $x_{i_{des}}$ and the measured joint positions, $x_i$. 
\begin{figure*}[!htb]
    \centering
    \begin{subfigure}[b]{0.30\textwidth}
        \includegraphics[trim=10 0 30 0,clip,width=\textwidth]{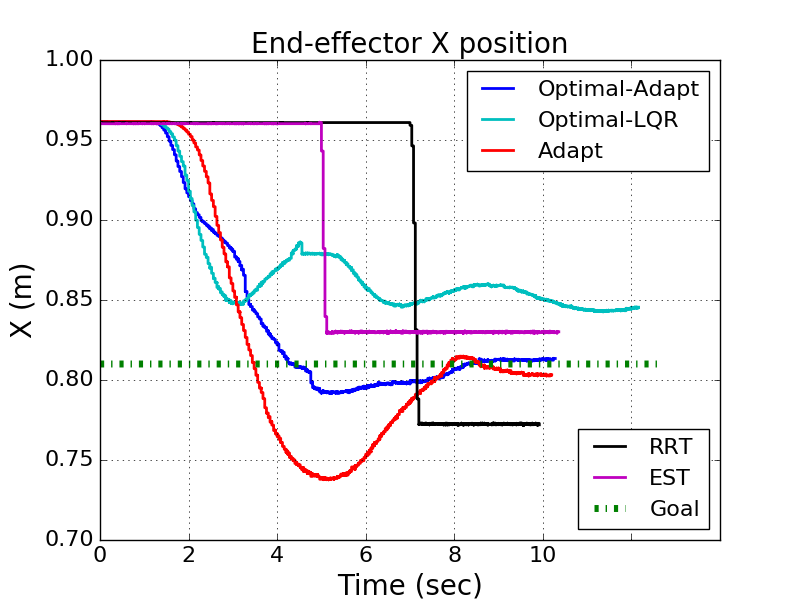}
        \caption{X end-effector position}
        \label{fig:x_pos}
    \end{subfigure}
    ~ 
    \begin{subfigure}[b]{0.30\textwidth}
        \includegraphics[trim=10 10 10 10,clip,width=\textwidth]{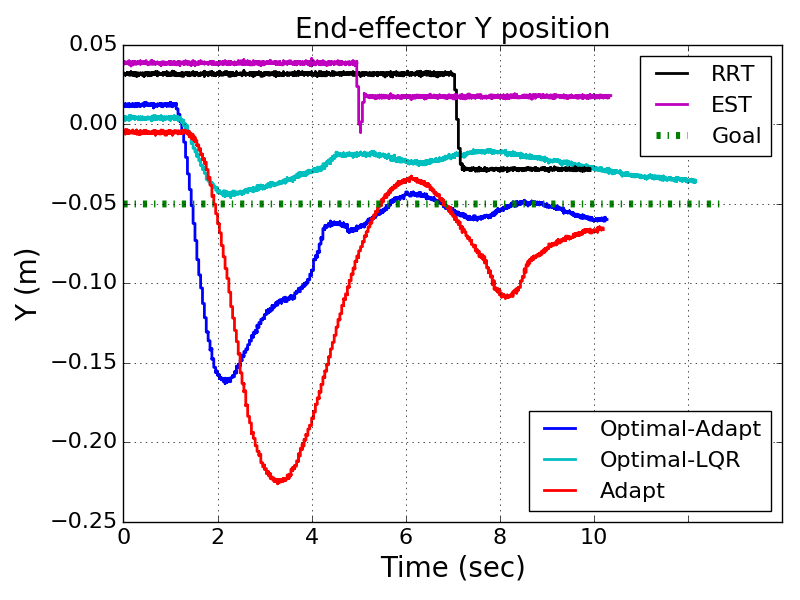}
        \caption{Y end-effector position}
        \label{fig:y_pos}
    \end{subfigure}
    ~ 
    \begin{subfigure}[b]{0.30\textwidth}
        \includegraphics[trim=10 10 10 0,clip,width=\textwidth]{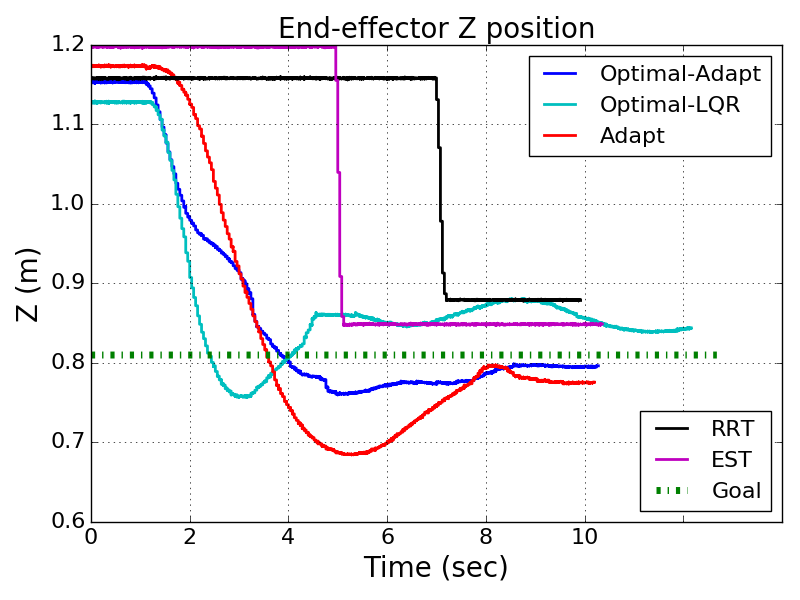}
        \caption{Z end-effector position}
        \label{fig:z_pos}
    \end{subfigure}
    \caption{\small{Comparison of end-effector behavior using Optimal-Adaptive method, Optimal-LQR method, Adaptive strategy, \ac{RRT} and \ac{EST} methods to reach a discrete point defined in world coordinates at $(0.81, -0.05, 0.8)$ meters.  The \ac{RRT} and \ac{EST} methods require more than $4$ seconds to compute a plan but have a very fast execution, while for the Adapt and Optimal-Adapt methods the execution  starts as soon as the first waypoint is generated, leading to a slower but more gradual movement. Using the Optimal-LQR method presents a higher variation from the set-point, showcasing the limitation of an inaccurate initial guess.}}\label{fig:ee_x_y_z}
\end{figure*}
\begin{figure*}[!htb]
  \centering
    \begin{subfigure}[]{\columnwidth}
    \centering
     \includegraphics[width=0.75\columnwidth]{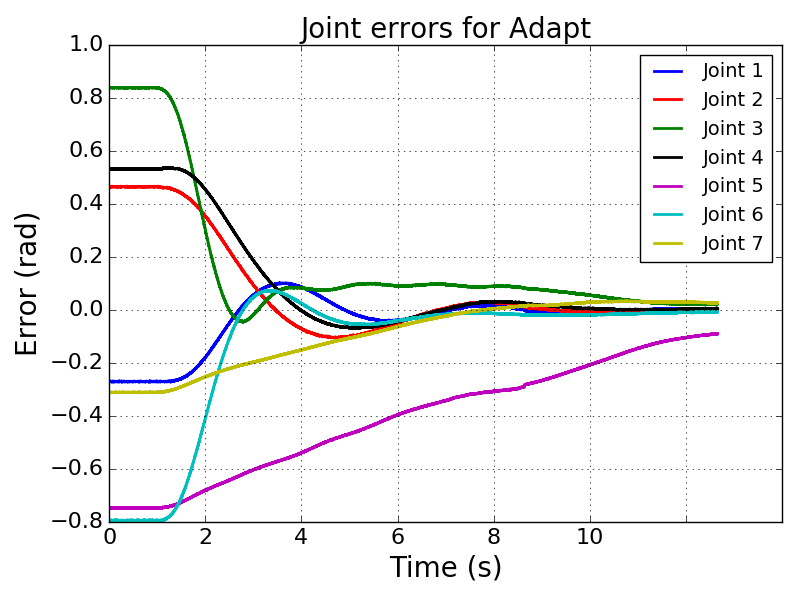}
     \caption{Adaptive joint errors}
     \label{fig:error_adapt}
     \end{subfigure}
 \begin{subfigure}[]{\columnwidth}
    \centering
     \includegraphics[width=0.75\columnwidth]{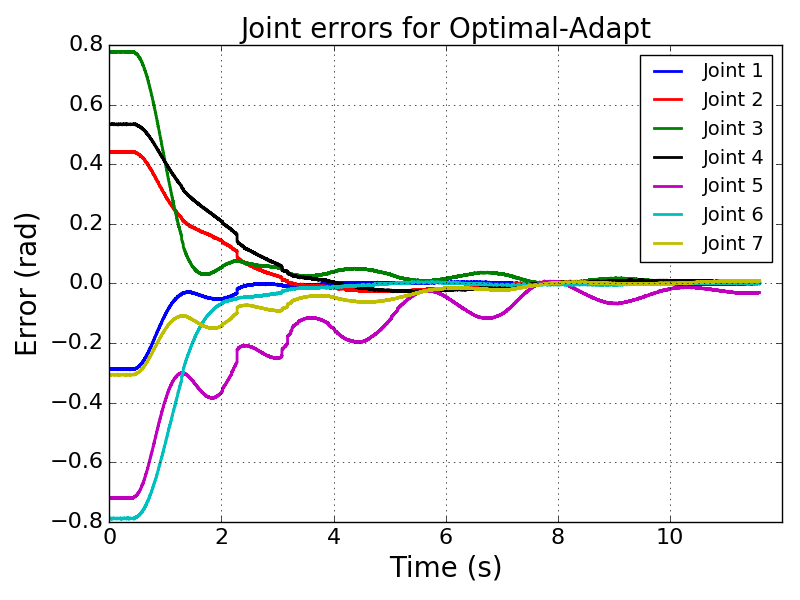}
     \caption{Optimal-Adaptive joint errors}
     \label{fig:error_optim}
     \end{subfigure}
     \caption{\small{Joint angle errors for the (a) initial Adaptive strategy and the (b) Optimal-Adaptive method when the end-effector behaviour is the one presented in Figure~\ref{fig:ee_x_y_z}. The Optimal-Adaptive strategy reduces significantly the time needed for the joints to reach the desired configuration compared with the initial Adaptive strategy, where also Joint $5$ is not able to reach the desired configuration, leading to an offset from the goal.}}
     \label{fig:joint_traj}
\end{figure*}

Incorporating Equations~(\ref{eq:null_space_reform})-(\ref{eq:feed_forward}) into the optimal formulation expressed by Equation (\ref{eq:problem_statement_final}), the optimal path that does not violate the constraints of the problem is defined as the solution of:
\begin{equation}
 \begin{aligned}
    \Delta u^{m*}_i =  \min_{\Delta u_i^m} &\lbrace  o_N + \Delta x_N^T q_N + \frac{1}{2} \Delta x_N^T Q_N + \Delta x_N  + \\
    & + \sum_{i=0}^{N-1} \tilde{o}_i + \Delta x_i \tilde{q}_i + \Delta {u_i^m}^T \tilde{r}_i + \frac{1}{2} \Delta x_i^T \tilde{Q}_i \Delta x_i +  \\
    & + \frac{1}{2} \Delta {u_i^m}^T \tilde{R}_i \Delta u_i^m + \Delta {u_i^m}^T \tilde{P}_i \Delta x_i  \rbrace
\end{aligned}
      \label{eq:problem_statement_reformulated_u}
  \end{equation}
  subject to
  \begin{equation}
    \Delta x_{i+1} = \tilde{A}_i \Delta x_i + \tilde{B}_i \Delta u_i^m + \tilde{k}_i, \quad x(0) =x_0
    \label{eq:dynamic_eq_reformulated_u}
\end{equation}
where
\begin{equation}
    \begin{aligned}
   \tilde{A}_i & = A_i + B_i \Gamma_i  \\
    \!
    \tilde{B}_i & = B_i \mathcal{N(\cdot)} \\
    \!
    \tilde{g}_i & = B_i \Theta_i \\
    \!
    \tilde{o}_i & = o_i +  \Theta_i^T r_i + \frac{1}{2} \Theta_i^T R_i \Theta_i \\
    \!
    \tilde{q}_i & = q_i + \Gamma^T r_i + P_i^T \Theta_i + \Gamma_i^TR_i \Theta_i \\
    \!
    \tilde{r} & = \mathcal{N}(\cdot) \left(r_i + R_i \Gamma_i \right) \\
    \!
    \tilde{Q}_i & = Q_i + \Gamma_i^T R_i \Gamma_i + \Gamma_i^T P_i + P_i^T \Gamma_i \\
    \!
    \tilde{R}_i & = \mathcal{N}(\cdot) R_i \mathcal{N}(\cdot) \\
    \!
    \tilde{P}_i & = \mathcal{N}(\cdot) \left(P_i + R_i \Gamma_i \right)
    \end{aligned}
    \label{eq:reparametrization}
\end{equation}

The solution of this optimal structure is computed using a Riccati equation formulation \cite{kuvcera1972discrete}, producing a feedback law presented in Equation (\ref{eq:optimal_ctrl_output}). 
\begin{equation}
\begin{aligned}
     \Delta u_i^m & = -\left( \tilde{R}_i + \tilde{B}_i^TS_{i+1}\tilde{B}_i\right)^{\dagger} \lbrace (\tilde{P}_i + \tilde{B}_i^TS_{i+1}\tilde{A}_i) \Delta x_i - \\
     & -\left[\tilde{r}_i + \tilde{B}_i^T(s_{i+1}+S_{i+1} k_i)\right] \rbrace
\end{aligned}
    \label{eq:optimal_ctrl_output}
\end{equation}
where $S_{i+1} \in \mathbb{R}^{n \times n}$ and $s_{i+1} \in \mathbb{R}^{n}$ are weighting matrices. 

The computational cost of the final planning solution, Equation (\ref{eq:final_ctrl_solution}), allows for  real-time implementation of this strategy on high-dimensional robotic systems, ensures the constraints of the problem are fulfilled and generates plans that can be fulfilled by the low-level controller. 
\begin{equation}
    \hat{u}_{i} = \hat{u}_{i_{des}} + \Delta u_i^c + \mathcal{N}(\cdot) \Delta u_i^m
    \label{eq:final_ctrl_solution}
\end{equation}
\begin{figure*}[!htb]
  \centering
    \begin{subfigure}[]{0.30\textwidth}
        \includegraphics[trim=10 0 10 0,clip, width=\textwidth]{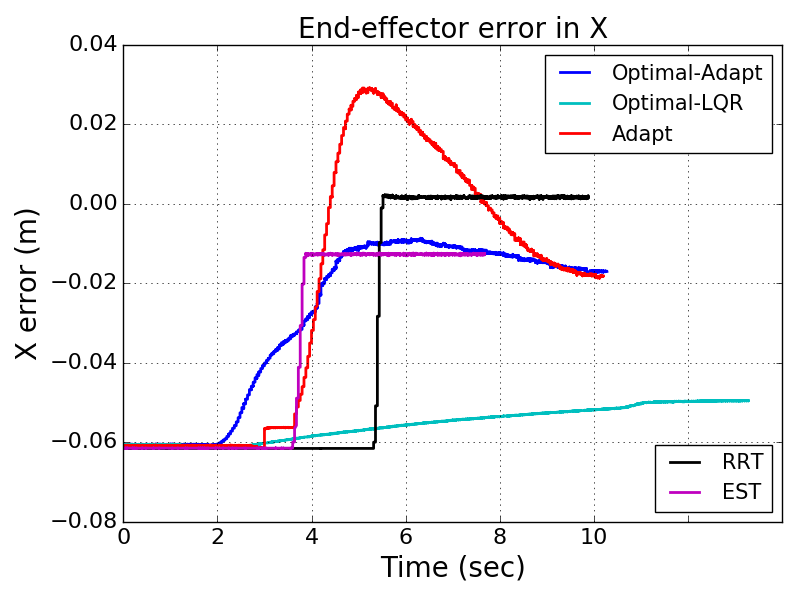}
         \caption{Error in X}
         \label{fig:err_x_lqr}
     \end{subfigure}
     ~
    \begin{subfigure}[]{0.30\textwidth}
        \includegraphics[trim=10 0 10 0,clip, width=\textwidth]{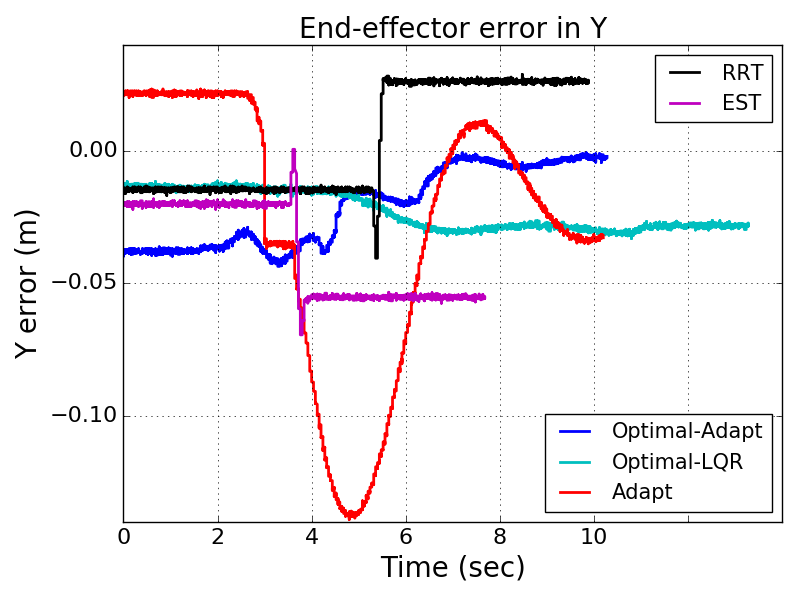}
        \caption{Error in Y}
        \label{fig:err_y_lqr}
    \end{subfigure}
    ~
    \begin{subfigure}[]{0.30\textwidth}
        \includegraphics[trim=10 0 10 0,clip, width=\textwidth]{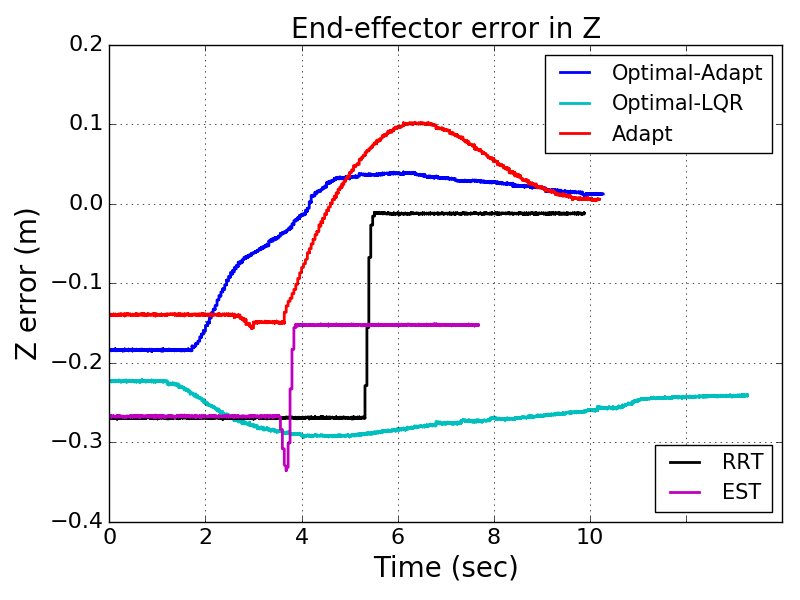}
        \caption{Error in Z}
        \label{fig:err_z_lqr}
    \end{subfigure}
    \caption{\small{End-effector error for the case when the joint limit presented in Figure~\ref{fig:joint_restricted} is imposed: the Optimal-Adapt (proposed) strategy reaches the vicinity of the goal fulfilling the constraints while the Optimal-LQR has a higher end-effector error, highlighting the importance of a reliable initial estimate. The \ac{RRT} method presents better accuracy  compared with any other method, nevertheless the constraints are broken and the \ac{EST} breaks the limits and also has a worse end-effector error.}\vspace{-10pt}}
     \label{fig:errors_ee}
\end{figure*}
\begin{figure}[!htb]
  \centering
    \includegraphics[width=0.75\columnwidth]{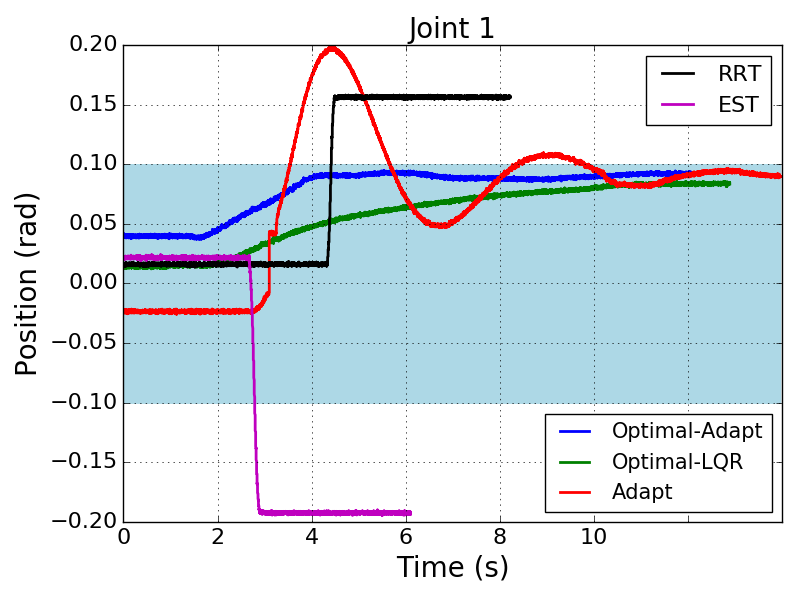}
     \caption{\small{Behaviour of Joint $1$ for the case when it can move only between ($-0.1$, $0.1$) radians. The proposed Optimal-Adapt strategy succeeds in maintaining the joint between the boundaries (in the shaded area of the graph) as well as the Optimal-LQR method, as they both rely on the same priority-optimal principle that handles constraints.}\vspace{-10pt}}
     \label{fig:joint_restricted}
\end{figure}

\vspace{-10pt}
\section{Experimental results}
\label{sec:results}
%
In this section we present the results of the proposed method applied on the $7$ \ac{DOF} manipulator of the Fetch~robot~\cite{wise2016fetch}, Figure~\ref{fig:fetch}.  
The strategy requires the manual tuning of $4$ parameters for each \ac{DOF}. A time horizon of $5$ seconds and a discretization step of $0.001$ seconds lead to an average of $6$ iterations until convergence and a CPU time of $0.008$ seconds $\left(\text{Intel(R) Core(TM)} \text{i}5-4570\text{S CPU} @ 2.90\text{GHz} \right)$. We showcase the behavior of the robot's manipulator using the proposed strategy considering both unconstrained and constrained environments.
The influence of the initial guess is discussed by analyzing the platform performance when using \textit{(i)} the initial adaptive estimation (noted as Adapt); \textit{(ii)} the proposed strategy (noted as Optimal-Adapt) and \textit{(iii)} a velocity \ac{LQR} initial estimation with the optimal strategy defined in Section \ref{sub:optimal} (noted as Optimal-LQR). Furthermore, we evaluate the Optimal-Adapt method in comparison with state-of-the-art  \ac{RRT} \cite{lavalle1998rapidly} and \ac{EST} \cite{phillips2004guided} implementations. 

%
\vspace{-10pt}
\subsection{Unconstrained environment}
In the first set of experiments we consider unconstrained movement for the robotic manipulator. Starting at an initial position the task objective is to reach a number of goals defined in world coordinates. 
In Table~\ref{tab:res_unconstrained}, we present a comparative evaluation between the Optimal-Adapt proposed strategy and the previously described baseline methods. 
%
We evaluate these algorithms over a set of $30$ different goals, in terms of \ac{RMSE} for the end-effector, success rate in finding a solution to reach the goal, and the mean and standard deviation of the time needed execute the action. Although the baseline algorithms have faster mean planning and execution times, this is highly dependent on the location of the goal. In the experiments we conducted, the planning times for these strategies fluctuated between $0.005$ seconds up to $20$ seconds, while our method remained between $6-8$ seconds for both execution and planning consistently, the fluctuation in time is visible in the \ac{STD} values in the table. The benefits of a robust model-based initial estimation in the proposed Optimal-Adapt strategy can be seen through the lower \ac{RMSE} and lower average time required to reach the goal compared with when the classical \ac{LQR} strategy is paired with the prioritized-method. 
\begin{table}[!htbp]
\centering
\begin{tabular}{|c|c|c|c|c|}
\hline \hline
Method & \ac{RMSE} (m) & Completion ($\%$) & Time (s) & \ac{STD} (s) \\
\hline
Optimal-Adapt & $0.109$ & $91$& $7.226$ & $2.408$\\
Optimal-LQR &$0.153$ & $91$ &$11.812$ &$5.065$\\
Adapt & $0.156$ & $91$& $10.250$& $1.998$\\
\hline
EST & $0.223$ & $78$& $5.199$& $7.271$\\
RRT & $0.188$ & $66$ & $6.504$& $6.197$\\
\hline \hline
\end{tabular}
\caption{\small{The average root-mean-square error for the end-effector position, success rate average total time and time standard deviation (planning and execution) of reaching the desired goals over a set of  $30$ different end-effector goals.}\vspace{-10pt}}
\label{tab:res_unconstrained}
\end{table}
The proposed method, Optimal-Adapt, has a higher success rate and lower \ac{RMSE} error compared with the \ac{EST} or \ac{RRT} algorithms. This can also be seen from Figure~\ref{fig:ee_x_y_z} where the end-effector's position reachs the goal with higher accuracy. While \ac{RRT} and \ac{EST} take the initial period to plan a path and then rapidly execute it, our algorithm combines the two components leading to a less erratic path. We also present a comparison with the initial adaptive path estimation and in Figure~\ref{fig:ee_x_y_z} it can be seen that the overshoot in the initial path is considerably reduced by the optimal-adaptive strategy. By looking at the joint errors shown in Figure~\ref{fig:joint_traj} it can be seen that the initial adaptive estimation method produces a slower response compared  to the optimal-adaptive strategy, but this comes at the cost of more oscillatory behaviour for some of the joints.

%

%
\vspace{-10pt}
\subsection{Constrained environment}
\begin{figure}[!htb]
  \centering
    \includegraphics[width=0.70\columnwidth]{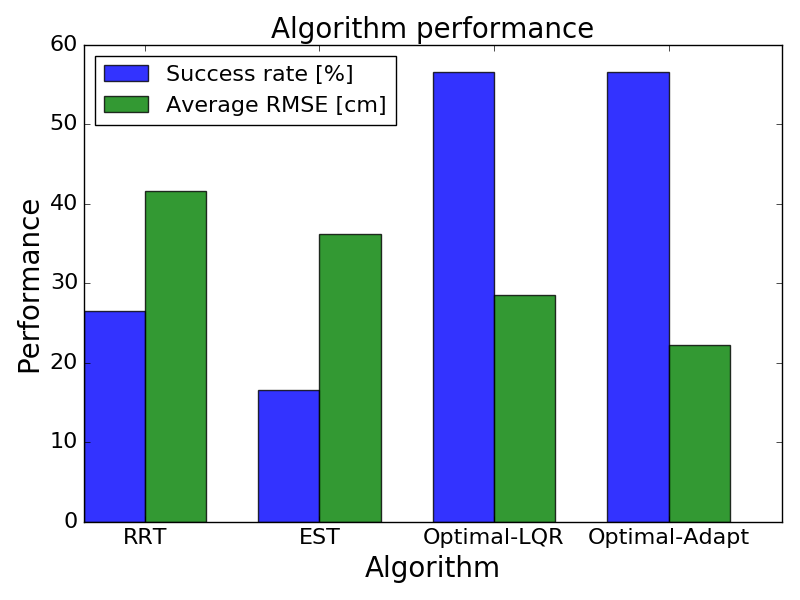}
     \caption{\small{\ac{RRT}, \ac{EST}, {Optimal-Adaptive} and {Optimal-LQR} performances in terms of number of plans that fulfilled the constraints (success rate) and average RMSE for those successful plan. The Optimal-Adaptive algorithm has higher successful rate and lower average RMSE over a set of $30$ different end-effector goals.}}
     \label{fig:alg_performance}
\end{figure}

In this example we illustrate the algorithm's performance when inequality constraints are applied to the system. We consider a scenario where the end-effector must reach a certain goal, but one of the joints is restricted in its movement. Considering that Joint $1$ can operate only between $-0.1$ and $0.1$ radians, in Figure~\ref{fig:joint_restricted} it can be seen that using both the optimal-adaptive strategy and the Optimal-LQR method, the platform is able to fulfill these restrictions, as both methods use the same the priority-optimization component. The initial adaptive path estimation is not considering constraints in its formulation and is not able to reach the goal without breaking the joint limits. When \ac{RRT} or \ac{EST} are considered, there is a trade-off between enforcing tighter joint limits or reaching the goal. From the end-effector position errors in Figure~\ref{fig:errors_ee} it can be seen that by fulfilling the constraints of the problem the optimal-adaptive strategy reaches very close proximity of the goal. The influence of a good initial estimate, using the Optimal-Adapt approach proposed in this paper, can be seen when comparing the end-effector position error with that of the Optimal-LQR strategy. In this case, although the joint limits are still maintained in the allowed range, the final location of the end-effector presents a higher offset from the desired location. The Optimal-Adapt strategy has better accuracy compared with \ac{RRT} or \ac{EST} methods as can be seen from Figure~\ref{fig:alg_performance}, where the algorithm performance is evaluated. We used a set of different $30$ end-effector goals, where for each experiment we limited the movement of one of the joints. The full proposed strategy Optimal-Adapt has a better success rate generating a viable path in the presence of enforced joint limitations compared with state-of-the-art methods. For the paths generated the end-effector to goal error is considerably lower for the proposed strategy, although compared with the case when no constraints are imposed to the system (in Table~\ref{tab:res_unconstrained}) the distance from the goal has increased. The Optimal-LQR strategy presents similar results in terms of planning and execution, while having an average end-effector error of $28.6$ centimeters compared with $22.3$ centimeters when the proposed Optimal-Adapt strategy is used. 

\vspace{-20pt}


\section{Conclusions}
\label{sec:conclusions}
In this paper we presented a new planning method subject to inequality constraints for redundant manipulators. The strategy uses a robust initial estimation strategy in a prioritized optimization framework. The initial path estimation is based on an adaptive low-level control law and an estimated dynamic model, providing a feasible path in open environments. This path is optimized to provide the best available solution and to take into account constraints providing a real-time solution to the problem. The strategy is validated with a $7$ \ac{DOF} manipulator and the results are comparable with state-of-the-art planning methods. The method is advantageous especially for cases when robotic systems have to work in cluttered and dynamic environments. 

\renewcommand{\bibfont}{\normalfont\small}
\printbibliography
\end{document}